\documentclass[10pt, a4paper]{article}

\usepackage[]{lrec-coling2024}
\usepackage{amsmath}

\title{Emotion Classification in Low and Moderate Resource Languages}

\name{Shabnam Tafreshi\textsuperscript{1} Shubham Vatsal\textsuperscript{2} Mona Diab\textsuperscript{3}} 

\address{
\textsuperscript{1}University of Maryland, Science Academy, Maryland, USA \\
\textsuperscript{2}New York University, CIMS, New York, USA \\
\textsuperscript{3}Carnegie Mellon University, LTI,  Pennsylvania, USA \\}

\abstract{
It is important to be able to analyze the emotional state of people around the globe. There are 7100+ active languages spoken around the world and building emotion classification for each language is labor intensive. Particularly for low-resource and endangered languages, building emotion classification can be quite challenging. We present a cross-lingual emotion classifier, where we train an emotion classifier with resource-rich languages (i.e. \textit{English} in our work) and transfer the learning to low and moderate resource languages. We compare and contrast two approaches of transfer learning from a high-resource language to a low or moderate-resource language. One approach projects the annotation from a high-resource language to low and moderate-resource language in parallel corpora and the other one uses direct transfer from high-resource language to the other languages. We show the efficacy of our approaches on 6 languages: \textit{Farsi}, \textit{Arabic}, \textit{Spanish}, \textit{Ilocano}, \textit{Odia}, and \textit{Azerbaijani}. Our results indicate that our approaches outperform random baselines and transfer emotions across languages successfully. For all languages, the direct cross-lingual transfer of emotion yields better results. We also create annotated emotion-labeled resources for four languages: \textit{Farsi}, \textit{Azerbaijani}, \textit{Ilocano} and \textit{Odia}.
 \\ \newline  }

\begin{document}

\maketitleabstract

\section{Introduction}
\label{intro}
Human sentiments and thoughts are mostly governed by their emotions. Human emotions can help us understand human behavior in different circumstances like natural disasters, civil unrest, or a pandemic and can give us insights into mental health. They can further help us in analyzing various social and cultural factors. Therefore, it is important to be able to analyze human emotions around the globe. However, obtaining data for every language that is spoken by people, is challenging, and in some cases when the language is not digitized, is impossible. Hence, it is important to design innovative ways to be able to analyze emotions for low and moderate-resource languages. This work is focused on universal as well as cross-cultural emotion models that have been studied by a number of psychologists and have been well adopted by Natural Language Processing (NLP) literature. 
The first research question we consider is which emotion tags or labels can be universal if any. The popular psychological emotion model \cite{ekman1993facial} which is commonly adopted by NLP tasks uses 6 basic emotions and most of the datasets in \textit{English} and other languages are annotated with this model. These 6 emotions include \textit{anger}, \textit{disgust}, \textit{fear}, \textit{happiness}, \textit{sadness} and \textit{surprise}. Based on the availability of data to evaluate our models, we choose a subset of these 6 emotions which includes \textit{anger}, \textit{fear}, and \textit{joy}.

In this work, we discuss two methods or approaches aiming at transferring emotions from resource-rich languages (source language i.e. \textit{English} in our study) to low and moderate resource languages (target language). The first approach uses a parallel in-genre or out-of-genre corpora to project annotations from the source language to the target language. In the second method, there are no parallel corpora available, and the emotions are directly transferred from \textit{source} language to the \textit{target} languages. The overall contribution of this study are as follows:
\begin{itemize}
    \item {Discussing and evaluating methods to transfer emotion from resource-rich language (i.e. \textit{English}) to low and moderate resource languages, even when digital resources are not available for the language.}
    \item {Studying the feasibility of transfer of emotion across languages by building context features to present emotion cues across languages.}
    \item {Collecting and annotating two emotion corpora in \textit{Ilocano}, \textit{Odia}, \textit{Farsi} and \textit{Azerbaijani} as novel resources for low resource and endangered languages}.
    \item {Emotion lexicon annotation for \textit{Ilocano} and \textit{Odia}.}
\end{itemize}
\section{Resource Collection and Annotation}
\label{rescollann}
\footnote{Please email the authors for getting access to the resources discussed in this work}This work is focused on building cross-lingual emotion models for 3 basic emotions \textit{anger}, \textit{fear}, and \textit{joy}. We evaluate our methods in 6 target languages: \textit{Farsi}, \textit{Arabic}, \textit{Spanish}, \textit{Ilocano}, \textit{Odia} and \textit{Azerbaijani} as spoken by a minority community in Iran. These languages are from different language families, typologies, and genres. They are also different in the level of their corresponding resource availability. \textit{Arabic} and \textit{Spanish} are moderate resource languages. \textit{Ilocano}, \textit{Odia}, and \textit{Farsi} are low resource languages and \textit{Azerbaijani} is an endangered language. Details about the collection and annotation of these languages are explained in the following subsections. Data statistics are shown in Table \ref{langdetails}.
\subsection{Farsi and Azerbaijani}
\label{prazLang}
\textit{Farsi} or \textit{Persian} \footnote{\textit{Farsi} is a pluricentric language predominantly spoken and used officially within Iran, Afghanistan, and Tajikistan in three mutually intelligible standard varieties, namely \textit{Iranian Persian}, \textit{Dari Persian} (officially named \textit{Dari} since 1958) \cite{olesen2013islam} and \textit{Tajiki Persian} (officially named \textit{Tajik} since the Soviet era) \cite{baker2019routledge}.} is a Western Iranian language belonging to the Iranian branch of the Indo-Iranian subdivision of the Indo-European languages.  \textit{Azerbaijani} or \textit{Azeri} also known as \textit{Azerbaijani Turkic} or \textit{Azerbaijani Turkish} is a Turkic language spoken primarily by the Azerbaijani people who live mainly in the Republic of Azerbaijan where the North Azerbaijani variety is spoken and in Iranian Azerbaijan where the South Azerbaijani variety is spoken.\footnote{\url{https://en.wikipedia.org/wiki/Azerbaijani_language}} This collection is from Iranian Azerbaijani which is written in Persian script and often mixed with Persian words.
\textit{Farsi} corpus leveraging emotion keywords for pattern matching and popular emojis is collected from Twitter over the span of two months, July and August of 2018. An intense cleaning process is done on this collection to be able to get a slice of this data ready for annotation. Tweets with less than 6 tokens are removed and the association of emojis with emotion tags is studied resulting in retention of tweets with only certain emojis. Using distant supervision, the data is annotated using Plutchik 8 basic emotions (PL8) \cite{plutchik1984emotions}. A classifier is trained with this noisy data and 10 fold cross-validation method is used to tag all the data. Then from these machine-labeled data, a very small collection, 800 tweets, with high confidence scores or softmax probabilities are selected for manual annotation. Annotators are graduate students with an NLP focus, knowledgeable for this particular task, and native speakers of Farsi. A set of instructions with sufficient explanation about the task and examples per emotion tag are provided. Each tweet is tagged by 3 annotators. Fleiss's Kappa \cite{fleiss1971measuring} is calculated for overall inter-annotator agreement and is reported to be 0.42\%. A total of 800 tweets are annotated from which emotions \textit{anger}, \textit{fear} and \textit{joy} are used for this study.

We collect \textit{Azerbaijani} data from native speakers. \textit{Azerbaijani} is considered an endangered language with no digital resource available for it. For collecting the data, we adopt the protocol described in prior work \citep{scherer1994evidence} where annotators are asked to describe their emotional experiences in text form. The emotion categories are PL8. Our annotators describe their emotions in sentence form for these basic emotion categories. We leverage two native \textit{Azeri} annotators for this task and they both cross-annotate each other's sentences. Kappa is calculated for overall inter-annotator agreement and is reported to be 0.32\%. 
\subsection{Ilocano and Odia}
\label{iloodia}
\textit{Ilocano} is from an Austronesian language family spoken in the Philippines. \textit{Odia} is from the Indo-Aryan language family spoken in the Indian state of Odisha. \textit{Odia} appears to have relatively little influence from \textit{Persian} and \textit{Arabic}. These languages are considered low-resource languages. \textit{Ilocano} and \textit{Odia} are collected and annotated by Appen \footnote{\url{https://appen.com}} which is a platform that provides data to companies and academics for their machine learning projects. This collection has in-genre, in-domain, and out-of-genre monolingual as well as parallel (\textit{English} being the other pair) data. Genres for these two languages are a mixture of news, discussion forums, and tweets. 
\subsection{Arabic and Spanish}
\textit{Arabic} is a Semitic language spoken primarily across the Arab world. \textit{Arabic} is morphologically rich and it has both derivative and inflectional morphology. \textit{Spanish} is a Romance language of the Indo-European language family that evolved from colloquial Latin spoken on the Iberian Peninsula of Europe. Both of these languages are among moderate resource languages. Both \textit{Arabic} and \textit{Spanish} data are taken from SemEval 2018 task collection \cite{mohammad2018semeval}.

\subsection{Religious Text (Bible \& Quran)}
\label{rtbq}
Religious text is available in parallel form and can be used as a source to train machine learning models or for extracting dictionaries or utilized as a source of training data for pre-trained multi-lingual resources (e.g. multi-lingual embedding). It is an unusual choice of a resource for emotion analysis as the genre, domain, and vocabulary are quite different from the choice of vocabulary, genre, and sentence structure that we often see in emotion analysis. However, such corpora have the advantage of being available in scale in many languages. We have used the Bible corpus of \cite{christodouloupoulos2015massively} which contains Bible translations for 100 languages and the Tanzil translations for Quran to create a combined parallel corpus. We only used these parallel corpora for \textit{Arabic}, \textit{Spanish} and \textit{Farsi} because \textit{Ilocano} and \textit{Odia} already have parallel corpora as we described above in section 2.3 and \textit{Azerbaijani} does not have translated Quran or Bible available digitally. \footnote{Azerbaijani that is spoken in Iran has a different script than Azerbaijani that is spoken in the Republic of Azerbaijan, hence in written form, there is no cross-usability with regards to language resources.}
\begin{table}[h!]
\begin{center}

\scalebox{0.9}{
\begin{tabular}{cccc}
      \hline
      \textbf{Language} & \textbf{Anger} & \textbf{Fear} & \textbf{Joy} \\
      \hline
      Arabic & 374 & 373 & 449 \\
      Spanish & 628 & 618 & 730 \\
      Ilocano & 24 & 12 & 83  \\
      Odia & 63 & 64 & 200 \\
      Persian (Farsi) & 15 & 50 & 148 \\
      Azerbaijani & 24 & 36 & 50 \\
\hline
\end{tabular}}
\caption{Number of Emotion Labels Per Language}
\label{langdetails}
\end{center}
\end{table}
\section{Annotation Projection in Emotion Classification}
\label{annpro}
For many years, researchers have experimented with the usage of parallel corpora for multilingual and cross-lingual NLP tasks. Text classification is one of those standard NLP tasks that can exploit parallel corpora to relieve the effort involved in creating annotations for new languages. However, the question that arises is how an NLP task like text classification exploits parallel corpora for multilingual and cross-lingual purposes. One of the ways to do so is by annotation projection. Annotation projection is the method to transfer linguistic annotations from resource-rich languages (e.g. \textit{English}) to low-resource languages.

In our study, we do single-source annotation projection where the only source language we use is \textit{English}. We chose \textit{English} as the source language because it is the most abundant language present out there in terms of various resources including labeled corpus, word embeddings etc. There are multiple assumptions associated with annotation projection using parallel corpora. One of the most important assumptions is that in the parallel corpora, every piece of text (word, sentence, document) of the source language is aligned with the target language. The next assumption is that there is an abundance of labeled data available in the source language which can be used for training machine learning models. We describe this monolingual emotion detection machine learning model (en\_mono) for \textit{English} which is our source language in annotation projection in section \ref{annproexp}. The annotation projection method steps are listed below.
\begin{itemize}
\item{En\_mono is used to label the English portion of the parallel corpora.}
\item{The emotion labels for the English portion of parallel corpora are projected to the target language.}
\item{A new emotion detection machine learning model (target\_mono) is trained using the newly annotated target language portion of parallel corpora from the previous step.}
\item{Target\_mono is used to predict labels of the test or held-out target language corpus.}
\end{itemize}

We use different parallel corpora for different target languages. We use the Bible and Quran for \textit{Arabic}, \textit{Spanish}, and \textit{Persian} as discussed in section \ref{rtbq} \footnote{We use Bible text in this study due to low results obtained from using Quran}. Parallel corpora for \textit{Ilocano} and \textit{Odia} are provided by Appen as explained in section \ref{iloodia}. For \textit{Azerbaijani}, there are no parallel corpora available. The key idea here is to test how much emotion is transferred from out-of-genre or out-of-domain corpora, and how much emotion cues can be transferred through in-genre or in-domain, but noisy corpora.
\subsection{Experiments}
\label{annproexp}
One of the objectives of this study is to investigate an efficient combination of different genres or sources to build sufficient training sources that can be robust toward unseen data. Empirical results from these studies \cite{tafreshi2018emotion,tafreshi2018sentence} suggest that a combination of news, blog posts, and semi-noisy tweets can create robust emotion models. There have been experiments with other combinations such as large amounts of noisy tweets with a small number of blog posts and news, however, the results suggest that a balanced combination of formal and non-formal text can suit best to create a robust model. Therefore, to label the \textit{English} portion of parallel corpora, the following corpora combination is used to train a classifier in \textit{English} (source language). For each corpus, we have listed down the corresponding distribution of data across our three chosen emotions.
\begin{itemize}
\item{Goodnews \cite{bostan2019goodnewseveryone}: \textit{anger} (885), \textit{fear} (419) and \textit{joy} (264)}
\item{Blog Posts \cite{aman2007identifying}: \textit{anger} (149), \textit{fear} (91) and \textit{joy} (479)}
\item{Saif15
\cite{mohammad2015using}: \textit{anger} (1118), \textit{fear} (1938) and \textit{joy} (5775)}
\end{itemize}

\begin{table}[h!]
\begin{center}
\scalebox{0.9}{
\begin{tabular}{ccc}
      \hline
      \textbf{Language} & \textbf{\# of Data Points} & \textbf{Data Source} \\
      \hline
      Arabic & 12679 & Bible \\
      Spanish & 12679 & Bible \\
      Ilocano & 12457 & Mix Genre \\
      {Odia} & 11654 & Mix Genre \\
      Persian (Farsi) & 12679 & Bible \\
      Azerbaijani & - & - \\
\hline
\end{tabular}}
\end{center}
\caption{Parallel Corpora Statistic}
\label{parallelcorpora}
\end{table}

Statistics of parallel corpora are illustrated in Table \ref{parallelcorpora}. All parallel corpora have an equal number of data points across all 3 emotion labels.

\begin{figure}[!ht]
\begin{center}
\includegraphics[scale=0.55]{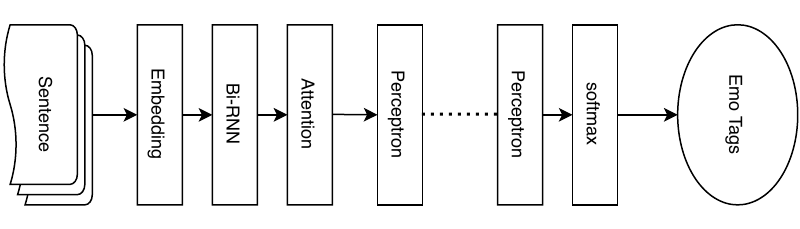} 
\end{center}
\caption{Bi-RNN Attention Model}
\label{paralleltagger}
\end{figure}

We experiment with two different en\_mono models. We start with a random baseline model. The reason why we chose a random baseline is because a lot of previous works or other baselines include language-dependent factors which cannot be directly applied in our case as we are dealing with some low-resource and endangered languages. For example, we could not have used any recent context-sensitive language models as they do not include endangered languages like Azerbaijani. Next, we experiment with a bidirectional recurrent neural network (Bi-RNN Attention) model with an attention mechanism. The features for this model are represented by fastText \footnote{\url{https://github.com/facebookresearch/fastText}} embedding. The architecture of this Bi-RNN Attention model is shown in Figure \ref{paralleltagger}. The layers of this model include an embedding layer of 300 dimensions followed by 70 units of Bi-RNN. Then, we have 3 layers of perceptrons, each having 50 units and finally followed by a softmax layer. The batch size used for training is 32 with 0.1 dropouts and the number of epochs depends on whether there has been any change in loss in the last 3 epochs or not.
\begin {table*}[h!]
\begin {center}
\scalebox{0.9}{
\begin {tabular}{ccccccccc}
\hline
\textbf{Language} & \multicolumn{4}{c}{\textbf{Arabic}} & \multicolumn{4}{c}{\textbf{Spanish}} \\ 
\hline
 \textbf{Emotion} & Anger & Fear & Joy & W-avg & Anger & Fear & Joy & W-avg\\ 
\hline
\textbf{Random Baseline} & 0.31 & 0.29 & 0.30 & 0.30 & 0.33 & 0.30 & 0.33 & 0.32\\
\textbf{Bi-RNN-Attention} & 0.05 & 0.45 & 0.20 & \textbf{0.33} & 0.30 & 0.32 & 0.57 & \textbf{0.44}\\
\hline
\end {tabular}}
\end {center}
\caption{Annotation Projection Results (Arabic \& Spanish)}
\label{arsp}
\end {table*}
\begin {table*}[h!]
\begin {center}
\scalebox{0.9}{
\begin {tabular}{ccccccccccccc}
\hline
\textbf{Language} & \multicolumn{4}{c}{\textbf{Ilocano}} & \multicolumn{4}{c}{\textbf{Odia}} &\multicolumn{4}{c}{\textbf{Persian (Farsi)}} \\ 
\hline
 \textbf{Emotion} & Anger & Fear & Joy & W-avg & Anger & Fear & Joy & W-avg & Anger & Fear & Joy & W-avg\\ 
\hline
\textbf{Random Baseline} & 0.31 & 0.18 & 0.46 & 0.40 & 0.29 & 0.23 & 0.45 & 0.38 & 0.42 & 0.05 & 0.21 & \textbf{0.35}\\
\textbf{Bi-RNN-Attention} & 0.22 & 0.15 & 0.74 & \textbf{0.57} & 0.21 & 0.25 & 0.57 & \textbf{0.51} & 0.26 & 0.11 & 0.45 & 0.30\\
\hline
\end {tabular}}
\end {center}
\caption{Annotation Projection Results (Ilocano, Odia \& Persian)}
\label{failoodi}
\end {table*}

\subsection{Results and Discussion}
\label{annproresdis}
Learning emotion cues within a language is a challenging task but when these cues are shared across different languages, it becomes even more challenging. In the annotation projection method, errors can propagate through the pipeline from the first step which is using labeled data in the source language (i.e. \textit{English} in our case) to label the parallel corpora which are usually out-of-genre. Other types of error can happen because of the assumption that text in parallel corpora across source and target language is aligned and thus will have the same emotion label which may not be the case always. To somewhat address the first error, it is crucial to have a system in the source language that can generalize emotion detection which in our approach is achieved by augmentation of data from different genres including news blog posts and tweets. The second error is harder to catch and certainly needs native speakers in the loop.

The results for the annotation projection method across different discussed target languages are shown in Table \ref{arsp} and \ref{failoodi}. We calculate \textit{f1-score} for all the emotion labels across the discussed languages and models. The parallel corpora for \textit{Farsi}, \textit{Spanish}, and \textit{Arabic} are out-of-genre which is the reason for relatively low results in these three languages compared to \textit{Ilocano} and \textit{Odia} which have partly in-genre parallel corpora. The Bi-RNN Attention model outperforms the random baseline model. The results indicate that the combination of several genres during the training of a model for the source language plays an important role when it comes to the robustness of the model.
\section{Direct Cross-Lingual Transfer of Emotions}
\label{dclingtemo}
Cross-lingual emotion classification aims to employ labeled data from the source language to label the data in the target language which has no annotations initially to train a classifier.
In the previous approach, we explain the process of how to select multiple resources and genres to combine together to create a robust emotion model. In this method, we focus more on how to represent features in a unified manner. We do embedding alignment so that the features of the language pair can be represented in the same cross-lingual embedding vector space. We discuss two different ways of embedding alignment where one is at the word level and the other one is at the sentence level and further compare their results. In addition, the emotion lexicon is used to present emotion features at the word level. We discuss them more in detail in section \ref{bi-ling-fasttext} and section \ref{laser}. Finally, deep neural network models are used to learn these feature representations, and emotion is transferred from the source language \textit{English} to target languages which include \textit{Arabic}, \textit{Spanish}, \textit{Ilocano}, \textit{Odia}, \textit{Persian} and \textit{Azerbaijani}. We assume here that low-resource languages have no training or development dataset to train a machine learning model and all the labeled data is used as the test set.
\begin{table*}[ht]
\begin{center}
\scalebox{0.9}{
\begin{tabular}{cccccc}
      \hline
      \textbf{Language} & \textbf{Arabic} & \textbf{Spanish} & \textbf{Ilocano} & \textbf{Odia} & \textbf{Persian (Farsi)} \\
      \hline
       \# of Tokens & 610605 & 40280 & 42163 & 25709 & 420042 \\
\hline
\end{tabular}}
\end{center}
\caption{Number of Tokens in Monolingual Word Embedding per Language}
\label{embdtat}
\end{table*}

\begin {table*}
\begin {center}
\scalebox{0.8}{
\begin {tabular}{ccccccccccccc}
\hline
\textbf{Language} & \multicolumn{4}{c}{\textbf{Arabic}} & \multicolumn{4}{c}{\textbf{Spanish}} &\multicolumn{4}{c}{\textbf{Persian (Farsi)}} \\ 
\hline
 \textbf{Emotion} & Anger & Fear & Joy & W-avg & Anger & Fear & Joy & W-avg & Anger & Fear & Joy & w-avg\\ 
\hline
\textbf{Random Baseline} & 0.31 & 0.29 & 0.30 & 0.30 & 0.33 & 0.30 & 0.33 & 0.32 & 0.42 & 0.05 & 0.21 & 0.35\\
\textbf{Cross-Ling-fastText} & 0.12 & 0.23 & 0.52 & 0.37 & 0.03 & 0.38 & 0.49 & 0.38 & 0.70 & 0.14 & 0.43 & 0.61\\
\textbf{Cross-Ling-fastText-af24} & 0.21 & 0.26 & 0.51 & 0.38 & 0.02 & 0.39 & 0.51 & 0.39 & 0.42 & 0.10 & 0.42 & 0.40\\
\textbf{Cross-Ling-LASER} & 0.43 & 0.66 & 0.75 & \textbf{0.62} & 0.61 & 0.60 & 0.63 & \textbf{0.61} & 0.56 & 0.66 & 0.73 & \textbf{0.67}\\
\hline
\end {tabular}}
\end {center}
\caption{Cross-lingual Emotion Results (Arabic, Spanish, Farsi)}
\label{clarspfa}
\end {table*}

\subsection{Cross-Lingual Word Embedding Alignment}
\label{bi-ling-fasttext}
In this embedding alignment, in order to create a unified cross-lingual vector space, we use pre-trained monolingual vector spaces (i.e. fastText) for the language pair and apply an unsupervised approach Multilingual Unsupervised and Supervised Embedding (MUSE) for the alignment. 
This method does not need any bilingual dictionary or parallel corpora to align the two monolingual vector spaces. Hence, it is suited for low-resource languages where usually there is no ground truth bilingual dictionary or sufficient parallel corpora available. Monolingual embedding for the source language (\textit{English}) and all 5 target languages, but \textit{Azerbaijani} are available and have been trained on their corresponding Wikipedias. 
\textit{Azerbaijani} does not have publicly available unstructured text and hence we could not build unified cross-lingual embedding for this language. To be able to generate aligned embedding for \textit{Azerbaijani}, we pivot to a language with more available resources. The script of \textit{Azerbaijani} is the same as that of \textit{Persian} and there are some overlaps at the word level. We got help from a native speaker of this language to mark the emotional words that are mentioned as the causality of the emotional context in each sentence. Then, these words are replaced in the \textit{Azerbaijani} corpus with Persian (Farsi) words. Finally, Persian-English-aligned embedding is used to predict emotion labels in \textit{Azerbaijani}. 

We also experiment with emotion lexicon as features to the aligned embedding achieved by this method. We use a 24-dimension (\textit{af24}) feature set for this purpose. It represents the distribution of emotional words and their intensity level in the context. This feature has 24 dimensions and each dimension presents one emotion from PL8's basic emotion categories in 3 intensity levels: high, medium, and low. To design this feature, we utilize the NRC emotion intensity lexicon dictionary \cite{mohammad-turney-2010-emotions}. This dictionary presents PL8's emotion categories with intensity levels of high, medium, or low for each entry word in the dictionary. Emotional words per sentence are extracted using this dictionary and this feature is populated accordingly with 1 for presence and 0 for absence of the emotion category. In the case of multiple entries per word in the dictionary, the emotion tag or label is selected based on the frequency of the word associated with a particular emotion label in the training set. The following example demonstrates the design of this feature.

\textit{I ran during the sunrise, it was the most \textbf{joyful} thing ever.}

In the lexicon dictionary, the word \textit{joyful} as present in the above sentence has emotion label \textit{joy} with high intensity and hence the dimension corresponding to this feature is marked 1 whereas every other dimension is represented by 0. NRC lexicon in English is manually annotated but the \textit{Arabic}, \textit{Spanish} and \textit{Persian} versions of this lexicon are machine translated. As a result, some of the emotion words are noisy.

NRC is not translated for \textit{Ilocano} and \textit{Odia}. We use the parallel corpora provided by Appen to create bi-lingual dictionaries for both \textit{Ilocano} and \textit{Odia}. We further use GIZA++ \citep{och2003systematic} to align the parallel corpora. Then these dictionaries are cleaned up and the intersection of these two dictionaries and NRC English lexicon dictionary are extracted as the emotion lexicon for both of these languages. 

\subsection{Direct Language-Agnostic SEntence Representations (LASER)}
\label{laser}
LASER\footnote{\url{https://github.com/facebookresearch/LASER}} is a library that is used for multilingual sentence embedding and it is available for 93 languages. We don't need to do any form of embedding alignment for LASER as the embedding for different languages is already aligned and thus they can be directly used. 
We use LASER sentence embedding available for 3 languages discussed in our study which includes \textit{Arabic}, \textit{Spanish}, and \textit{Farsi}. 
\subsection{Experiments}
\label{crossexp}
We experiment with four different models for \textit{Arabic}, \textit{Spanish}, and \textit{Persian}. We start with a random baseline model. Then we have a Bi-RNN Attention model where we use an aligned cross-lingual embedding with monolingual embedding coming from fastText as discussed in section \ref{bi-ling-fasttext}. The Bi-RNN Attention model architecture is exactly the same as that of what we discussed in section \ref{annproexp} and is shown in Figure \ref{paralleltagger}. We refer to this model by Cross-Ling-fastText. Next, we have an extension of our previous Bi-RNN model where apart from having the aligned cross-lingual embedding, we also have the \textit{af24} feature set. This model goes by the name of Cross-Ling-fastText-af24. Finally, we have another Bi-RNN model that uses LASER sentence embedding as discussed in section \ref{laser}. This model is called Cross-Ling-LASER.

We experiment with three different models for \textit{Ilocano} and \textit{Odia} whereas only two different models for \textit{Azerbaijani}. We again start with a random baseline followed by Cross-Ling-fastText and Cross-Ling-fastText-af24 for \textit{Ilocano} and \textit{Odia}. For \textit{Azerbaijani}, we experiment with random baseline and Cross-Ling-fastText-af24. The reason why we have Cross-Ling-fastText-af24 for \textit{Azerbaijani} but not Cross-Ling-fastText is because we don't have fastText embedding for \textit{Azerbaijani} but we identify emotion words in \textit{Azerbaijani} corpus and replace them by \textit{Persian} words as discussed in section \ref{bi-ling-fasttext}.

Again, the reason why we chose a random baseline for all the languages is because a lot of previous works include language-dependent factors which cannot be directly applied in our case as we are dealing with some low-resource and endangered languages. For example, recent context-sensitive language models do not include endangered languages like Azerbaijani.

\begin {table*}
\begin {center}

\scalebox{0.8}{
\begin {tabular}{ccccccccccccc}
\hline
\textbf{Language} & \multicolumn{4}{c}{\textbf{Ilocano}} & \multicolumn{4}{c}{\textbf{Odia}} &\multicolumn{4}{c}{\textbf{Azerbaijani}} \\ 
\hline
 \textbf{Emotion} & Anger & Fear & Joy & W-avg & Anger & Fear & Joy & W-avg & Anger & Fear & Joy & W-avg\\ 
\hline
\textbf{Random Baseline} & 0.31 & 0.18 & 0.46 & 0.40 & 0.29 & 0.23 & 0.45 & 0.38 & 0.21 & 0.26 & 0.44 & 0.33\\
\textbf{Cross-Ling-fastText} & 0.20 & 0.10 & 0.68 & 0.53& 0.26 & 0.22 & 0.73 & \textbf{0.54} & - & - & - & -\\
\textbf{Cross-Ling-fastText-af24} & 0.24 & 0.15 & 0.72 & \textbf{0.58} & 0.06 & 0.16 & 0.71 & 0.47 & 0.11 & 0.48 &  0.34 & \textbf{0.41}\\
\hline
\end {tabular}}
\end {center}
\caption{Cross-lingual Emotion Results (Ilocano, Odia, Azerbaijani)}
\label{cliloodiaaz}
\end {table*}

\begin {table*}
\begin {center}
\scalebox{0.9}{
\begin {tabular}{ccccccc}
\hline
\textbf{Method} & \multicolumn{6}{c}{\textbf{Language}} \\ 
\hline
 & Arabic & Spanish & Persian & Ilocano & Odia & Azerbaijani\\ 
\hline
\textbf{Annotation Projection}             & 0.33 & 0.44 & 0.30 & 0.57 & 0.51 & -\\
\textbf{Cross-Ling-fastText}               & 0.37 & 0.38 & 0.61 & 0.53 & \textbf{0.54} & -\\
\textbf{Cross-Ling-fastText-af24}          & 0.38 & 0.39 & 0.40 & \textbf{0.58} & 0.47 & \textbf{0.41}\\
\textbf{Cross-Ling-LASER}                  & \textbf{0.62} & \textbf{0.61} & \textbf{0.67} & - & - & -\\
\hline
\end {tabular}}
\end {center}
\caption{Annotation Projection and Cross-lingual Overall Results}
\label{resultcom}
\end {table*}
\subsection{Results and Discussion}
The results of direct cross-lingual transfer of emotions models are illustrated in Table \ref{clarspfa} and \ref{cliloodiaaz}. We calculate \textit{f1-score}
for all the emotion labels across discussed languages and models. For \textit{Arabic}, \textit{Spanish} and \textit{Farsi}, Cross-Ling-LASER outperforms other models. This performance can be attributed to the fact that the contextual cues (clauses, phrases, and words) are embedded in the sentence in contrast to word embedding. An important point to discuss is the impact of \textit{af24} feature set on emotion categories in different languages. We note a positive impact for emotion labels \textit{anger} and \textit{fear} in \textit{Arabic}, \textit{joy} in \textit{Spanish} and \textit{anger}, \textit{fear} and \textit{joy} in Ilocano. In contrast, we notice a negative impact for all emotion labels in \textit{Farsi}. This is because most of the data points for Persian in our collection have long text with so many emotion words included in the context but a lot of emotion cues are unmarked. This causes \textit{af24} feature set to be noisy and hence creates a negative impact. There is also a negative impact of \textit{af24} for all emotion labels in \textit{Odia}. This is because the overlap between \textit{Odia} corpus and \textit{Odia} lexicon dictionary is very low (total of 30 words).

For \textit{Ilocano}, \textit{Odia} and \textit{Azerbaijani}, Cross-Ling-fastText and Cross-Ling-fastText-af24 models outperform the baseline.

The main takeaways from these results can be summarized as:

\begin{itemize}
    \item{The feature set \textit{af24} has a positive impact on improving model performances for 4 languages and hence it is an important factor representing emotion in a low resource setting.}
    \item{Emotion lexicon dictionary constructed for \textit{Odia} and \textit{Ilocano} from parallel corpora can be used for other NLP tasks related to emotion.}
    \item{The approach exploring similar script with word level overlaps between \textit{Azerbaijani} and \textit{Persian} turns out to be highly effective for classification of emotion for an endangered low resource language like \textit{Azerbaijani}.}
\end{itemize}

\section{Overall Discussion and Error Analysis}
Table \ref{resultcom} illustrates the overall results that are achieved by the two approaches and their variations in this study. Results indicate that overlap-genre or in-genre training creates better results in emotion classification. Direct cross-lingual transfer of emotion has better model performance compared to parallel annotation projection which is out-of-genre training for 3 languages \textit{Arabic}, \textit{Spanish} and \textit{Persian} and \textit{overlap-genre} training for \textit{Ilocano} and \textit{Odia}. The addition of \textit{af24} feature set in the cross-lingual transfer of emotions positively impacts the model performance for 4 languages: \textit{Ilocano}, \textit{Azerbaijani}, \textit{Arabic} and \textit{Spanish}. 
Despite the fact that whether the annotation projection method involves out-of-genre or overlap-genre training, we can observe comparable results with respect to direct cross-lingual transfer fastText models in two languages \textit{Arabic} and \textit{Odia} and better results for \textit{Ilocano}.

We discuss some examples along with their English translations in Figure \ref{wmp} per language which none of the discussed models could predict correctly.
{\textit{Arabic} - Sentence corresponding to this language contains the word \textbf{like} (marked in green) which has the emotion label \textit{joy} in the dictionary, none of the models are able to predict the emotion label for this sentence correctly. 
{\textit{Persian (Farsi)} - There is no emotion word marked in this sentence, interestingly all cross-lingual models have the same wrong prediction, (\textit{joy}), this could either be the fault of the w2v model or model is biased towards a lexicon.
{\textit{Spanish} - In this sentence it is hard to imply if it is the fault of encoding or model bias. Although the word \textbf{resentful} in the dictionary has a label \textit{anger}. 
{\textit{Ilocano and Odia} - English translation for these two low-resource languages on a sentence level is not available. These 2 sentences have no sign of emotion words which could be one of the causes for errors in these two languages.
{\textit{Azerbaijani} - In this sentence, \textbf{angry} (marked in green) is code-switched with Farsi, and it is marked as a flag for \textit{anger}, however, the model could not predict this sentence correctly.

\begin{figure}[!ht]
\begin{center}
\includegraphics[scale=0.19]{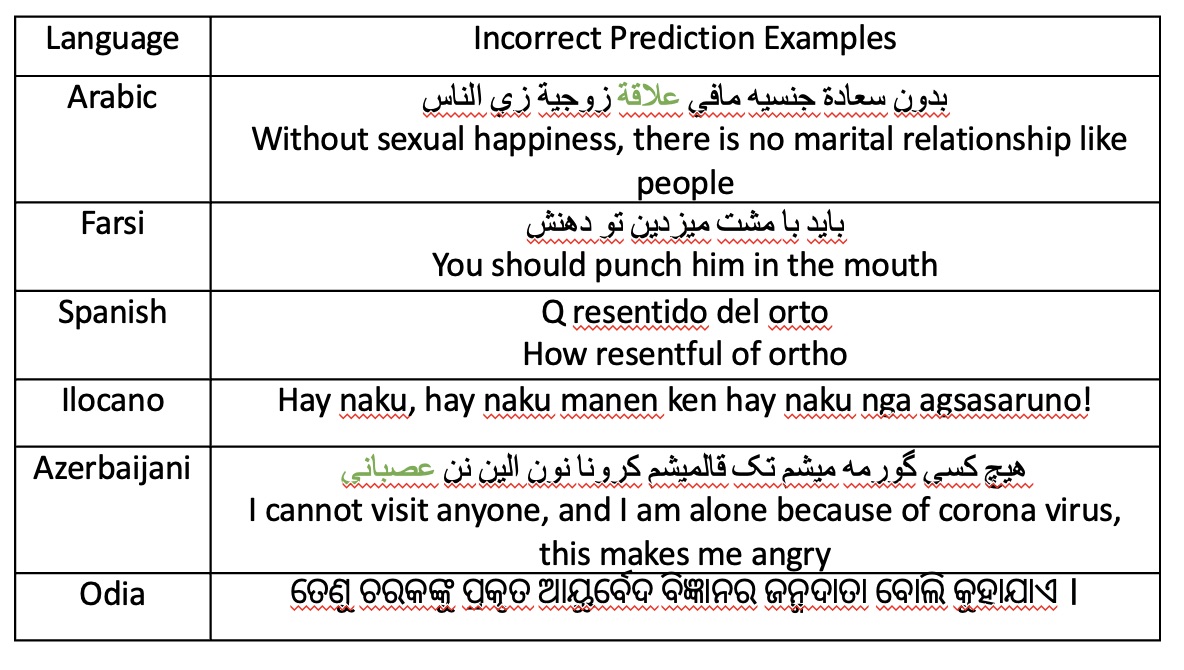}
\label{crosslingmodel1}
\end{center}
\caption{Wrong Model Predictions for All Languages}
\label{wmp}
\end{figure}
\section{Related Work}
\label{reword}

\textbf{Annotation projection} - these works \cite{mihalcea2007learning,rasooli2018cross,farra2019cross} explored this method for sentiment analysis. Mihalcea built a Romanian subjectivity classifier using annotation projections from a parallel English-Romanian corpus. Rasooli and Farra built a sentiment classifier for low-resource languages.
\textbf{Direct Transfer Learning} - If parallel corpora are available sentence representation can be learned from auto-encoders, once sentences are represented a machine learning model can learn sentiment in \textit{source} language and transfer to the \textit{target} language. One such work in \textit{English-Chinese} is a study by \cite{zhou2014bridging}. There are approaches that jointly train bilingual sentiment embeddings and the classifier in language pair, bilingual sentiment lexicon is used to minimize the projection matrices in language pair \cite{barnes2018bilingual}. The most related to this study in sentiment is the work by \cite{farra2019cross}. This work studies various techniques to create cross-lingual representations and we report the best results. These representations are in word level \cite{aldarmaki2018unsupervised,conneau2017word,berard2016multivec}. Current state-of-the-art cross-lingual models for emotions are pre-trained transformer language models such as Multilingual BERT (M-BERT) \cite{devlin2018bert} and XLM-RoBERTa \cite{conneau2019unsupervised}, however, currently, all these models are only available for a limited number of languages, hence, not the best candidate for low-resource languages.
\textbf{Zero- or few-shot Learning} - Large language models (LLMs) have gained popularity in the recent past for their effectiveness in reasoning where traditional computational methods failed \cite{thirunavukarasu2023large,vatsal2024survey}. But recent studies show that LLMs either lag behind supervised and dictionary-based learning or need to be combined with supervised learning in regard to their abilities to understand emotion \cite{zhao2023chatgpt, schick2020exploiting, min2022rethinking, barnes2023sentiment}. Unfortunately, none of these large language models are available for low-resource or moderate-resource languages. 
\section{Conclusion and Future Direction}
\label{conclusion}
In this paper, we study two broad approaches and many variations of these two approaches to analyze emotions in six low and moderate-resource languages without any machine translation. In one of the approaches, we focus on annotation projection by exploiting the available parallel corpora for the given languages. In the other approach, we talk about embedding alignment and how using an additional emotion lexicon feature set can boost emotion classification performance. We also collect and annotate two emotion corpora in \textit{Farsi}, \textit{Azerbaijani}, \textit{Ilocano} and \textit{Odia} as novel resources for low-resource and endangered languages. We further also build an emotion lexicon for \textit{Ilocano} and \textit{Odia}.
As a part of future work, we can expand our work to more low-resource languages and include more emotions that are valid across languages. Causation of emotion has been loosely studied in this work, yet the results are very promising and we believe cross-lingual emotion causality produces more meaningful results in both cross-lingual and monolingual emotion classification.
\section{Limitation}
In this paper, we only explored embedding on the word level to represent the low-resource languages and sentence representation which explores the representation of textual features in context. The reason for the latter choice is to show how contextual models work in moderate resource languages. Further improvement is possible if different pre-trained models are used.
We only explored 3 different emotion categories while it is possible to analyze more emotion categories across the languages. 
Exploring large language models remains a big limitation in this work, however, these approaches are interesting and have great potential in our future research direction.

\nocite{*}
\section{Bibliographical References}\label{sec:reference}

\bibliographystyle{lrec-coling2024-natbib}
\bibliography{lrec-coling2024-example}

\begin{thebibliography}{33}
\expandafter\ifx\csname natexlab\endcsname\relax\def\natexlab#1{#1}\fi

\bibitem[{Aldarmaki et~al.(2018)Aldarmaki, Mohan, and
  Diab}]{aldarmaki2018unsupervised}
Hanan Aldarmaki, Mahesh Mohan, and Mona Diab. 2018.
\newblock Unsupervised word mapping using structural similarities in
  monolingual embeddings.
\newblock \emph{Transactions of the Association of Computational Linguistics},
  6:185--196.

\bibitem[{Aman and Szpakowicz(2007)}]{aman2007identifying}
Saima Aman and Stan Szpakowicz. 2007.
\newblock Identifying expressions of emotion in text.
\newblock In \emph{International Conference on Text, Speech and Dialogue},
  pages 196--205. Springer.

\bibitem[{Baker and Saldanha(2019)}]{baker2019routledge}
Mona Baker and Gabriela Saldanha. 2019.
\newblock \emph{Routledge encyclopedia of translation studies}.
\newblock Routledge.

\bibitem[{Barnes(2023)}]{barnes2023sentiment}
Jeremy Barnes. 2023.
\newblock Sentiment and emotion classification in low-resource settings.
\newblock In \emph{Proceedings of the 13th Workshop on Computational Approaches
  to Subjectivity, Sentiment, \& Social Media Analysis}, pages 290--304.

\bibitem[{Barnes et~al.(2018)Barnes, Klinger, and Walde}]{barnes2018bilingual}
Jeremy Barnes, Roman Klinger, and Sabine Schulte~im Walde. 2018.
\newblock Bilingual sentiment embeddings: Joint projection of sentiment across
  languages.
\newblock \emph{arXiv preprint arXiv:1805.09016}.

\bibitem[{B{\'e}rard et~al.(2016)B{\'e}rard, Servan, Pietquin, and
  Besacier}]{berard2016multivec}
Alexandre B{\'e}rard, Christophe Servan, Olivier Pietquin, and Laurent
  Besacier. 2016.
\newblock Multivec: a multilingual and multilevel representation learning
  toolkit for nlp.

\bibitem[{Bostan et~al.(2019)Bostan, Kim, and
  Klinger}]{bostan2019goodnewseveryone}
Laura Bostan, Evgeny Kim, and Roman Klinger. 2019.
\newblock Goodnewseveryone: A corpus of news headlines annotated with emotions,
  semantic roles, and reader perception.
\newblock \emph{arXiv preprint arXiv:1912.03184}.

\bibitem[{Christodouloupoulos and
  Steedman(2015)}]{christodouloupoulos2015massively}
Christos Christodouloupoulos and Mark Steedman. 2015.
\newblock A massively parallel corpus: the bible in 100 languages.
\newblock \emph{Language resources and evaluation}, 49(2):375--395.

\bibitem[{Conneau et~al.(2019)Conneau, Khandelwal, Goyal, Chaudhary, Wenzek,
  Guzm{\'a}n, Grave, Ott, Zettlemoyer, and Stoyanov}]{conneau2019unsupervised}
Alexis Conneau, Kartikay Khandelwal, Naman Goyal, Vishrav Chaudhary, Guillaume
  Wenzek, Francisco Guzm{\'a}n, Edouard Grave, Myle Ott, Luke Zettlemoyer, and
  Veselin Stoyanov. 2019.
\newblock Unsupervised cross-lingual representation learning at scale.
\newblock \emph{arXiv preprint arXiv:1911.02116}.

\bibitem[{Conneau et~al.(2017)Conneau, Lample, Ranzato, Denoyer, and
  J{\'e}gou}]{conneau2017word}
Alexis Conneau, Guillaume Lample, Marc'Aurelio Ranzato, Ludovic Denoyer, and
  Herv{\'e} J{\'e}gou. 2017.
\newblock Word translation without parallel data.
\newblock \emph{arXiv preprint arXiv:1710.04087}.

\bibitem[{Devlin et~al.(2018)Devlin, Chang, Lee, and
  Toutanova}]{devlin2018bert}
Jacob Devlin, Ming-Wei Chang, Kenton Lee, and Kristina Toutanova. 2018.
\newblock Bert: Pre-training of deep bidirectional transformers for language
  understanding.
\newblock \emph{arXiv preprint arXiv:1810.04805}.

\bibitem[{Ekman(1993)}]{ekman1993facial}
Paul Ekman. 1993.
\newblock Facial expression and emotion.
\newblock \emph{American psychologist}, 48(4):384.

\bibitem[{Farra(2019)}]{farra2019cross}
Noura Farra. 2019.
\newblock \emph{Cross-Lingual and Low-Resource Sentiment Analysis}.
\newblock Ph.D. thesis, Columbia University.

\bibitem[{Fleiss(1971)}]{fleiss1971measuring}
Joseph~L Fleiss. 1971.
\newblock Measuring nominal scale agreement among many raters.
\newblock \emph{Psychological bulletin}, 76(5):378.

\bibitem[{Mihalcea et~al.(2007)Mihalcea, Banea, and
  Wiebe}]{mihalcea2007learning}
Rada Mihalcea, Carmen Banea, and Janyce Wiebe. 2007.
\newblock Learning multilingual subjective language via cross-lingual
  projections.
\newblock In \emph{Proceedings of the 45th annual meeting of the association of
  computational linguistics}, pages 976--983.

\bibitem[{Min et~al.(2022)Min, Lyu, Holtzman, Artetxe, Lewis, Hajishirzi, and
  Zettlemoyer}]{min2022rethinking}
Sewon Min, Xinxi Lyu, Ari Holtzman, Mikel Artetxe, Mike Lewis, Hannaneh
  Hajishirzi, and Luke Zettlemoyer. 2022.
\newblock Rethinking the role of demonstrations: What makes in-context learning
  work?
\newblock \emph{arXiv preprint arXiv:2202.12837}.

\bibitem[{Mohammad et~al.(2018)Mohammad, Bravo-Marquez, Salameh, and
  Kiritchenko}]{mohammad2018semeval}
Saif Mohammad, Felipe Bravo-Marquez, Mohammad Salameh, and Svetlana
  Kiritchenko. 2018.
\newblock Semeval-2018 task 1: Affect in tweets.
\newblock In \emph{Proceedings of The 12th International Workshop on Semantic
  Evaluation}, pages 1--17.

\bibitem[{Mohammad and Turney(2010)}]{mohammad-turney-2010-emotions}
Saif Mohammad and Peter Turney. 2010.
\newblock \href {https://aclanthology.org/W10-0204} {Emotions evoked by common
  words and phrases: Using {M}echanical {T}urk to create an emotion lexicon}.
\newblock In \emph{Proceedings of the {NAACL} {HLT} 2010 Workshop on
  Computational Approaches to Analysis and Generation of Emotion in Text},
  pages 26--34, Los Angeles, CA. Association for Computational Linguistics.

\bibitem[{Mohammad(2018)}]{LREC18-AIL}
Saif~M. Mohammad. 2018.
\newblock Word affect intensities.
\newblock In \emph{Proceedings of the 11th Edition of the Language Resources
  and Evaluation Conference (LREC-2018)}, Miyazaki, Japan.

\bibitem[{Mohammad and Kiritchenko(2015)}]{mohammad2015using}
Saif~M Mohammad and Svetlana Kiritchenko. 2015.
\newblock Using hashtags to capture fine emotion categories from tweets.
\newblock \emph{Computational Intelligence}, 31(2):301--326.

\bibitem[{Och and Ney(2003)}]{och2003systematic}
Franz~Josef Och and Hermann Ney. 2003.
\newblock A systematic comparison of various statistical alignment models.
\newblock \emph{Computational linguistics}, 29(1):19--51.

\bibitem[{Olesen(2013)}]{olesen2013islam}
Asta Olesen. 2013.
\newblock \emph{Islam \& Politics Afghanistan N}.
\newblock Routledge.

\bibitem[{Plutchik(1984)}]{plutchik1984emotions}
Robert Plutchik. 1984.
\newblock Emotions: A general psychoevolutionary theory.
\newblock \emph{Approaches to emotion}, 1984:197--219.

\bibitem[{Rasooli et~al.(2018)Rasooli, Farra, Radeva, Yu, and
  McKeown}]{rasooli2018cross}
Mohammad~Sadegh Rasooli, Noura Farra, Axinia Radeva, Tao Yu, and Kathleen
  McKeown. 2018.
\newblock Cross-lingual sentiment transfer with limited resources.
\newblock \emph{Machine Translation}, 32(1-2):143--165.

\bibitem[{Scherer and Wallbott(1994)}]{scherer1994evidence}
Klaus~R Scherer and Harald~G Wallbott. 1994.
\newblock Evidence for universality and cultural variation of differential
  emotion response patterning.
\newblock \emph{Journal of personality and social psychology}, 66(2):310.

\bibitem[{Schick and Sch{\"u}tze(2020)}]{schick2020exploiting}
Timo Schick and Hinrich Sch{\"u}tze. 2020.
\newblock Exploiting cloze questions for few shot text classification and
  natural language inference.
\newblock \emph{arXiv preprint arXiv:2001.07676}.

\bibitem[{Sun et~al.(2020)Sun, Ahn, Park, Tsvetkov, and
  Mortensen}]{sun2020cross}
Jimin Sun, Hwijeen Ahn, Chan~Young Park, Yulia Tsvetkov, and David~R Mortensen.
  2020.
\newblock Cross-cultural similarity features for cross-lingual transfer
  learning of pragmatically motivated tasks.
\newblock \emph{arXiv preprint arXiv:2006.09336}.

\bibitem[{Tafreshi and Diab(2018{\natexlab{a}})}]{tafreshi2018emotion}
Shabnam Tafreshi and Mona Diab. 2018{\natexlab{a}}.
\newblock Emotion detection and classification in a multigenre corpus with
  joint multi-task deep learning.
\newblock In \emph{Proceedings of the 27th international conference on
  computational linguistics}, pages 2905--2913.

\bibitem[{Tafreshi and Diab(2018{\natexlab{b}})}]{tafreshi2018sentence}
Shabnam Tafreshi and Mona Diab. 2018{\natexlab{b}}.
\newblock Sentence and clause level emotion annotation, detection, and
  classification in a multi-genre corpus.
\newblock In \emph{Proceedings of the Eleventh International Conference on
  Language Resources and Evaluation (LREC 2018)}.

\bibitem[{Thirunavukarasu et~al.(2023)Thirunavukarasu, Ting, Elangovan,
  Gutierrez, Tan, and Ting}]{thirunavukarasu2023large}
Arun~James Thirunavukarasu, Darren Shu~Jeng Ting, Kabilan Elangovan, Laura
  Gutierrez, Ting~Fang Tan, and Daniel Shu~Wei Ting. 2023.
\newblock Large language models in medicine.
\newblock \emph{Nature medicine}, 29(8):1930--1940.

\bibitem[{Vatsal and Dubey(2024)}]{vatsal2024survey}
Shubham Vatsal and Harsh Dubey. 2024.
\newblock A survey of prompt engineering methods in large language models for
  different nlp tasks.
\newblock \emph{arXiv preprint arXiv:2407.12994}.

\bibitem[{Zhao et~al.(2023)Zhao, Zhao, Lu, Wang, Tong, and
  Qin}]{zhao2023chatgpt}
Weixiang Zhao, Yanyan Zhao, Xin Lu, Shilong Wang, Yanpeng Tong, and Bing Qin.
  2023.
\newblock Is chatgpt equipped with emotional dialogue capabilities?
\newblock \emph{arXiv preprint arXiv:2304.09582}.

\bibitem[{Zhou et~al.(2014)Zhou, He, and Zhao}]{zhou2014bridging}
Guangyou Zhou, Tingting He, and Jun Zhao. 2014.
\newblock Bridging the language gap: Learning distributed semantics for
  cross-lingual sentiment classification.
\newblock In \emph{Natural Language Processing and Chinese Computing}, pages
  138--149. Springer.

\end{thebibliography}


\end{document}